\pdfoutput=1

\documentclass[11pt]{article}

\usepackage{latex/acl}

\usepackage{times}
\usepackage{latexsym}

\usepackage[T1]{fontenc}

\usepackage[utf8]{inputenc}

\usepackage{microtype}
\usepackage{fontawesome}
\usepackage{inconsolata}

\usepackage{hyperref}
\usepackage{url}
\usepackage{tikz}
\usepackage{multirow}
\usepackage{xspace}
\usepackage{booktabs}
\usepackage{ulem}
\usepackage{graphicx} 
\usepackage{highlight}
\usepackage{caption}
\usepackage{tcolorbox}
\usepackage{fancyvrb}
\usepackage{fvextra}
\usepackage{wrapfig}
\usepackage{bbm}
\usepackage[flushleft]{threeparttable}
\usepackage{enumitem}
\usepackage{verbatim}
\usepackage{scalerel} 

\usepackage{hyperref}

\usepackage{listings}
\usepackage{xcolor}

\definecolor{codegreen}{rgb}{0,0.6,0}
\definecolor{codegray}{rgb}{0.5,0.5,0.5}
\definecolor{codepurple}{rgb}{0.58,0,0.82}
\definecolor{backcolour}{rgb}{0.95,0.95,0.92}

\lstdefinestyle{mystyle}{
    backgroundcolor=\color{backcolour},   
    commentstyle=\color{codegreen},
    keywordstyle=\color{magenta},
    numberstyle=\tiny\color{codegray},
    stringstyle=\color{codepurple},
    basicstyle=\ttfamily\footnotesize,
    breakatwhitespace=false,         
    breaklines=true,                 
    captionpos=b,                    
    keepspaces=true,                 
    numbers=left,                    
    numbersep=5pt,                  
    showspaces=false,                
    showstringspaces=false,
    showtabs=false,                  
    tabsize=2
}

\title{PCA-Bench: Evaluating Multimodal Large Language Models in \\ Perception-Cognition-Action Chain}

\author{Liang Chen$^1$, Yichi Zhang$^1$, Shuhuai Ren$^1$, Haozhe Zhao$^1$, Zefan Cai$^1$, Yuchi Wang$^1$, \\ \textbf{Peiyi Wang$^1$, Xiangdi Meng$^1$, Tianyu Liu$^2$, Baobao Chang$^1$}\\
$^1$ National Key Laboratory for Multimedia Information Processing, Peking University\\$^2$ Alibaba Group \\
 \texttt{\{leo.liang.chen, yczhang, shuhuai\_ren\}@stu.pku.edu.cn}  \\
 \texttt{tianyu0421@alibaba-inc.com, chbb@pku.edu.cn} 
 \\ \faGithub \space \href{https://github.com/pkunlp-icler/PCA-EVAL}{\texttt{PCA-EVAL}}
 \includegraphics[height=10pt,width=10pt]{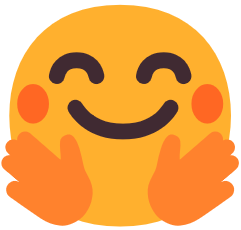} \href{https://huggingface.co/datasets/PCA-Bench/PCA-Bench-V1}{\texttt{PCA-Bench-V1}}}

\begin{document}
\maketitle
\begin{abstract}

We present PCA-Bench, a multimodal decision-making benchmark for evaluating the integrated capabilities of Multimodal Large Language Models (MLLMs). Departing from previous benchmarks focusing on simplistic tasks and individual model capability, PCA-Bench introduces three complex scenarios: autonomous driving, domestic robotics, and open-world games. Given task instructions and diverse contexts, the model is required to seamlessly integrate multiple capabilities of Perception, Cognition, and Action in a reasoning chain to make accurate decisions.  Moreover, PCA-Bench features error localization capabilities, scrutinizing model inaccuracies in areas such as perception, knowledge, or reasoning. This enhances the reliability of deploying MLLMs. To balance accuracy and efficiency in evaluation, we propose PCA-Eval, an automatic evaluation protocol, and assess 10 prevalent MLLMs. The results reveal significant performance disparities between open-source models and powerful proprietary models like GPT-4 Vision. To address this, we introduce Embodied-Instruction-Evolution (EIE), an automatic framework for synthesizing instruction tuning examples in multimodal embodied environments. EIE generates 7,510 training examples in PCA-Bench and enhances the performance of open-source MLLMs, occasionally surpassing GPT-4 Vision (+3\% in decision accuracy), thereby validating the effectiveness of EIE. Our findings suggest that robust MLLMs like GPT4-Vision show promise for decision-making in embodied agents, opening new avenues for MLLM research.

\end{abstract}

\section{Introduction}




\begin{figure}[!t]
    \centering
    \includegraphics[width=0.48\textwidth]{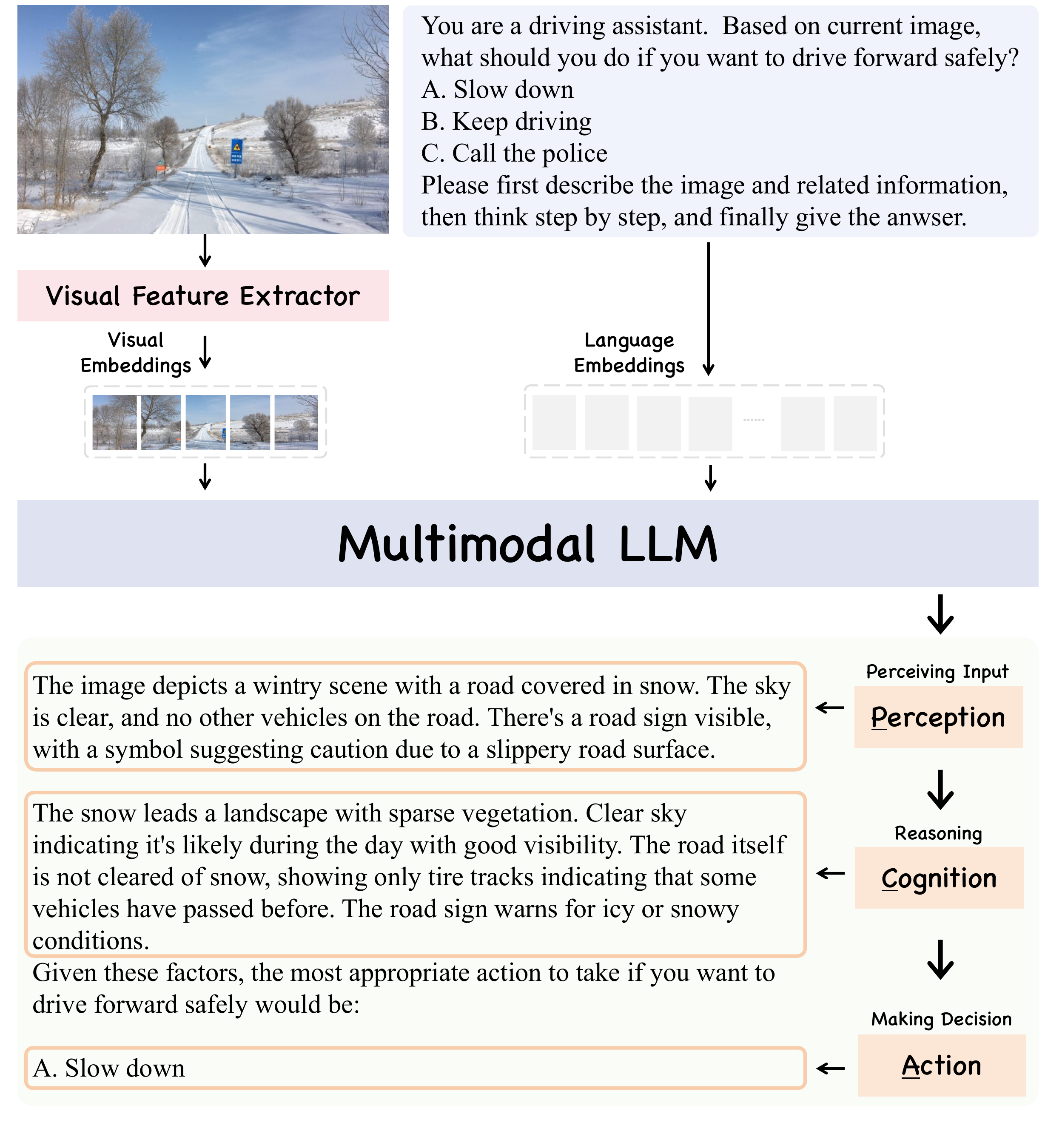}
    \caption{Example of decision making with MLLMs in the Perception-Cognition-Action Chain. 
    }
    \label{fig:pca-chain}
\end{figure}

Multimodal Large Language Models (MLLMs) have demonstrated remarkable capabilities in tackling complex tasks that necessitate a chain of integrated skills, including visual perception, world knowledge, reasoning, action, and more~\citep{gpt4v,dai2023instructblip,liu2023llava,li2023m3it,zhao2023mmicl}.

However, current MLLM benchmarks often evaluate these capabilities individually~\citep{fu2023mme,liu2023mmbench}, overlooking the significant integrated potential that Large Language Models (LLMs) contribute to multimodal models. While some benchmarks like MMMU~\citep{yue2023mmmu} and MathVista~\citep{Lu2023MathVistaEM} require abilities from both the vision and language part, they lack error localization techniques beyond accuracy assessments. This complicates identifying which part of the MLLM malfunctioned when making mistakes—whether it was the visual or the language component—and determines which aspect requires enhancement to enhance overall performance.

To address the challenges of insufficient integrated benchmarking and error localization problems, we introduce \textbf{PCA-Bench}. It arises with MLLM's applications in embodied AI and decision making, where models called agents need to first process multimodal observation from different environments, reason with the current situation and goal, and finally make an action from a given action space. The abilities in the complex decision making process can be abstracted to Perception, Cognition and Action according to the Perception-Action loop~\citep{FUSTER2004143} in Cognitive Science, a fundamental concept that describes how organisms process sensory information to interact with their environment through actions, offering a comprehensive framework for assessment. Figure~\ref{fig:pca-chain} shows how MLLMs make decisions in the PCA chain.

\begin{figure}[!t]
\centering
\includegraphics[width=0.5\textwidth]{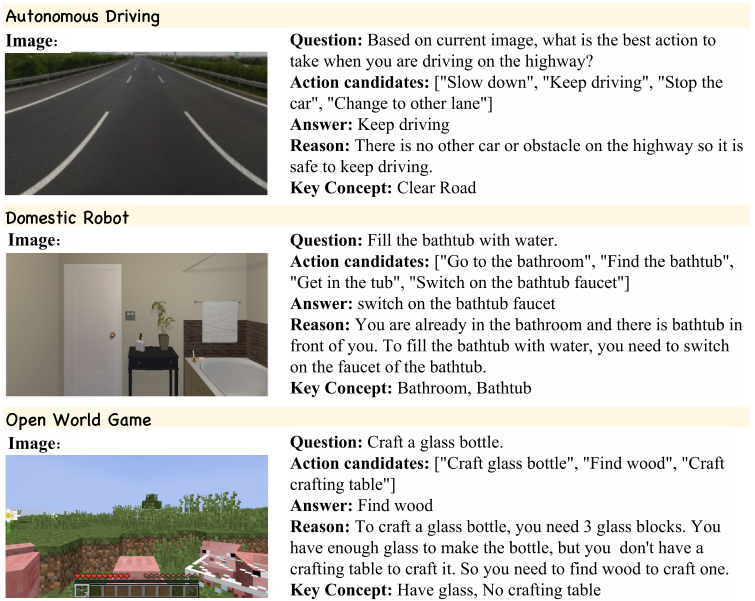}
  \caption{Instances of PCA-Bench in 3 domains.}
  \label{fig:example-pcaeval}
\end{figure}

The instances in PCA-Bench are from three influential domains in embodied decision-making: autonomous driving, domestic robotics, and open-world gaming. As shown in Figure~\ref{fig:example-pcaeval}, each instance is annotated by human annotators with a 6-element tuple: $<$\textit{image}, \textit{question}, \textit{action candidates}, \textit{answer}, \textit{reason}, \textit{key concept}$>$. The last three elements serve as anchors for error localization for Action, Cognition and Perception, correspondingly.


\textbf{PCA-Eval} is an anchor-based evaluation protocol, designed to automatically conduct error localization utilizing the powerful semantic parsing ability of LLMs and the anchor information in data annotation. In the past, such localization was both labor-intensive and time-consuming. PCA-Eval with strong LLMs like GPT4 demonstrates a strong kappa correlation with human assessments, reaching 0.8+ average kappa coefficients for perception, cognition, and action scores. The anchor-based evaluation provides the LLMs with groundtruth answers for each sub-score, preventing the systematic bias of LLM evaluators, such as position bias~\citep{wang2023large,zheng2023judging} in the pair-wise evaluation and verbosity bias~\citep{zheng2023judging} in simple preference evaluation. 
We also compared open state-of-the-art LLMs in PCA-Eval. Though they lag behind close ones in alignment with human assessments, we see large improvement when the model scales up. We believe that with specific training for error localization and improved general ability of open LLMs in the future, they would be more suitable evaluation tools for the reproducible and transparent characteristics.

Aiming at scaling up PCA-Bench, using LLM to synthesize training examples is an increasingly popular method for enhancing models without additional human involvement. We expand this approach to  generate more samples following the PCA guideline. Unlike text-based instruction generation methods like Self-Instruct~\citep{wang2023selfinstruct}, generating instructions in embodied environments poses distinct challenges. It demands not only the creation of textual instructions but also the generation of corresponding precise observations. To address these challenges, we propose \textbf{Embodied Instruction Evolution (EIE)}, which integrates external environments with LLMs, thereby extending the LLMs’ ability to data synthesize across various embodied environments, contributing to 7,510 training data in PCA-Bench.

We conduct comprehensive experiments and analysis on PCA-Bench, our findings are summarized as follows:

1. Visual perception and reasoning with world knowledge are two core abilities for an MLLM to make correct decisions in PCA-Bench. GPT4-Vision shows strong zero-shot cross-modal reasoning ability for embodied decision-making tasks, surpassing open-source MLLMs and even Tool-Using LLM-agent.

2. EIE could generate training samples significantly enhancing the performance of open-source MLLMs (surpassing GPT-4V at some scores), validating the effectiveness of the method.

3. PCA-Eval serves as a good error locator. Above the high average kappa coefficient (0.8+) with human assessments and its ability to pinpoint the error source, it can effectively distinguishes whether a model's correct decisions are fluky or through genuine understanding. This leads to a better ensemble metric for MLLM evaluation named Genuine PCA Score.

\section{PCA-Bench}
\subsection{Problem Definition}

Multimodal decision-making problems are commonly formalized with a partially observable Markov decision process. For MLLMs $\mathcal{F}$ tested in PCA-Bench, we care about given the multi-modal observation $o \in O$, the goal description $g$, a subset of candidates actions $A_C \subseteq A$, whether the model could make correct action $a \in A_C$ and give proper reasoning process $r$. 


\begin{equation}
\mathcal{F} ( g,o,A_C ) = (a,r)
\end{equation}


As shown in Figure~\ref{fig:example-pcaeval}, each instance in the benchmark is a 6-element tuple: $<$\textit{image}, \textit{question}, \textit{action candidates}, \textit{answer}, \textit{reason}, \textit{key concept}$>$. 
The image is collected from various embodied environments, including transportation scenes, housekeeper environments, and Minecraft. Questions, action candidates, and answers are derived from real tasks within the corresponding environment. The reasons explain why the answer is the best choice for the current image, while the key concept highlights the most question-related aspect of the image.

Unlike traditional visual question-answering datasets that emphasize visual perception (e.g., VQA~\citep{balanced_vqa_v2}) or visual reasoning (e.g., NLVR~\citep{Suhr2017NLVR}), PCA-Bench mandates accurate observation perception, complex task decomposition, and understanding the outcomes of various actions simultaneously.
Compared to embodied simulation environments such as ALFRED~\citep{ALFRED20} and Minedojo~\citep{fan2022minedojo}, PCA-Bench stands out for its focus on high-level actions, proving to be more effective for evaluating MLLMs. This is because high-level actions, which can be readily translated or programmed into low-level actions within their respective domains, are inherently more accessible to LLMs. The high-level actions are more comprehensible for LLMs than the direct low-level actions like action vectors in the simulation environments because (1) the high-level actions are in the form of natural languages, making it easier for LLMs to understand the meaning and connect with world knowledge. (2) LLMs are not grounded with low-level actions during the pretraining or finetuning stage, making it hard for LLMs to understand the consequences of executing an action.


To answer a question in PCA-Bench, the agent must possess the following abilities: (1) \textbf{Perception}: Accurately identify the concept related to the question within the image; (2) \textbf{Cognition}: Engage in reasoning based on image perception and worldly knowledge; (3) \textbf{Action}: Comprehend the potential actions, selecting the one that best aligns with the outcome of the reasoning process. A deficiency in any of these abilities would possibly result in an incorrect answer, posing a significant challenge to the more integrated capabilities of MLLMs. 


\subsection{PCA-Eval}

For each instance, we prompt the model to deliver an answer comprising a reasoning process $r$, and a final action $a$, represented as $<r, a>$. By comparing the model prediction with the ground truth answer, we can obtain a fine-grained diagnosis of the decision making process as follows:


\noindent \textbf{Perception Score (P-Score)} measures the model's accuracy in perceiving the observation. It is computed based on whether the agent's reasoning process $r$ includes the key concept of the instance. 
A score of 1 is assigned if at least one question-related key concept is described by the agent; otherwise, it is 0. 
For the top example in Figure~\ref{fig:example-pcaeval}, the agent should output ``clear road'' or ``no car visible'' or other semantically equivalent concepts in its description of the image to get the perception score. 

Parsing the model's output and determining whether it entails the key concept using shallow features of the sentence is not trivial. We leverage LLM to conduct entailment detection, which turns out to have a high alignment with human judgment. 

\noindent \textbf{Cognition Score (C-Score)} assesses the model's ability to reason, comprehend, and make informed decisions based on the perceived input data and world knowledge. The score is 1 if the reasoning process is correct, otherwise the score is 0. For the instance in Figure~\ref{fig:example-pcaeval}, the agent should link the ``clear road'' to the action ``keep driving'' based on transportation commonsense to get the score.

\noindent \textbf{Action Score (A-Score)} measures the model's ability to generate appropriate and effective responses or actions based on the perceived input data and the cognitive understanding of the context. The score is assigned a value of 1 if the agent selects the correct action; otherwise, the score is set to 0. 



\subsection{Automatic Evaluation}
\label{sec:pca-tool}

Recent advancements have seen researchers harnessing powerful LLMs for the evaluation of the output of language models. Studies have revealed that the outcomes from LLMs could exhibit remarkable alignment with human judgments \cite{zheng2023judging, wang2023large, wang2023making}. In our investigation, we employed GPT-4 to automatically evaluate perception, cognition, and action scores based on the model's outputs. Our findings underscore a significant agreement between GPT-4 scoring and human evaluation results. This is substantiated by Cohen-Kappa coefficients of 0.71, 0.82, and 0.94 for perception, cognition, and action evaluations, respectively. Experiments of human evaluation and comparison of open LLMs are in section~\ref{llm-eval}. For a detailed description of our evaluation tool, kindly refer to Appendix \ref{app:ae}.

\subsection{Benchmark Dataset Overview}

For the test set, the examples are written by 3 human experts for each domain. There are no overlapped environmental observations between the training and test sets. The details of the human annotation pipeline can be found in Appendix~\ref{app:human_anno_pipelines}. We introduce the three domains encompassed by our dataset as follows:

\paragraph{Autonomous Driving.} In the autonomous driving domain, instances are derived from real-world transportation scenes, which requires the agent to have particular abilities such as traffic sign recognition, obstacle detection, and decision-making at intersections. The dataset aims to evaluate an agent's ability to perceive and interpret visual information while making safe and efficient driving decisions. The images are collected from TT100K~\citep{Zhe_2016_CVPR_traffic_sign} dataset and annotators are instructed to propose an image-conditioned question that is grounded with real actions of vehicles. 

\paragraph{Domestic Robot.} The domestic assistance domain features instances from the ALFRED~\citep{ALFRED20,ai2thor} environment, which simulates a housekeeper robot performing tasks within a household setting. These tasks may include object manipulation, navigation, and interaction with various appliances. The environment assesses an agent's ability to understand and execute complex instructions while navigating and interacting with a dynamic environment. Annotators are asked to select one image from the randomly generated scenes in the environment, propose a question related to the items on the scene, and annotate the full information of the instance. 

\begin{figure}
  \begin{center}
    \includegraphics[width=0.4\textwidth]{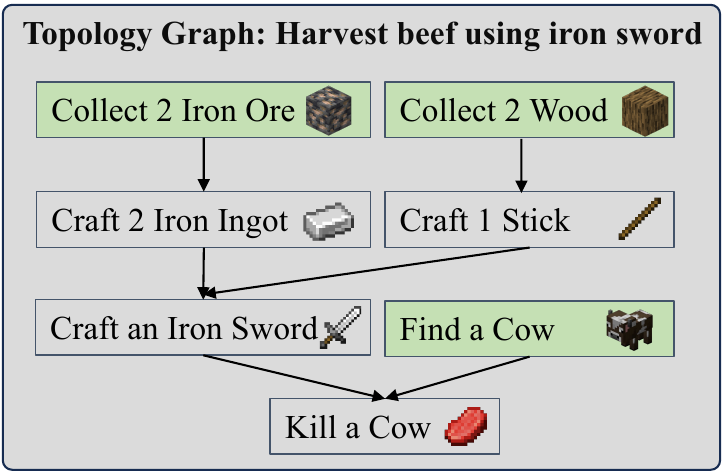}
  \end{center}
  \caption{Illustration of task topology graph. Events in green represent the leaf nodes of the graph.}
  \label{fig:topology}
\end{figure}

\paragraph{Open-World Game.} In the open-world game domain, instances are sourced from the Minecraft environment, where agents are tasked with exploring, crafting, and surviving in a procedurally generated world. This dataset evaluates an agent's ability to reason and plan actions within a complex, open-ended environment, which often requires long-term strategizing and adaptability. Annotators receive predefined tasks from MineDojo~\citep{fan2022minedojo} as a reference during the task generation phase. For each task, we instruct the annotator to sketch a task topology graph, exemplified in Figure~\ref{fig:topology}. The task should be completed under the topological order of the graph, where the event located in the leaf nodes should be finished first. Each node in the task topology graph can be viewed as a step in the sequential decision. We list the in-domain task distribution in Appendix~\ref{app:example_pca_eval}.

\subsection{Embodied Instruction Evolution}

The PCA-Bench benchmark also includes subset of automatic generated samples by Embodied Instruction Evolution(EIE), which is used as training set in our experiment. 

\begin{figure*}[!t]
    \centering
    \includegraphics[width=\textwidth]{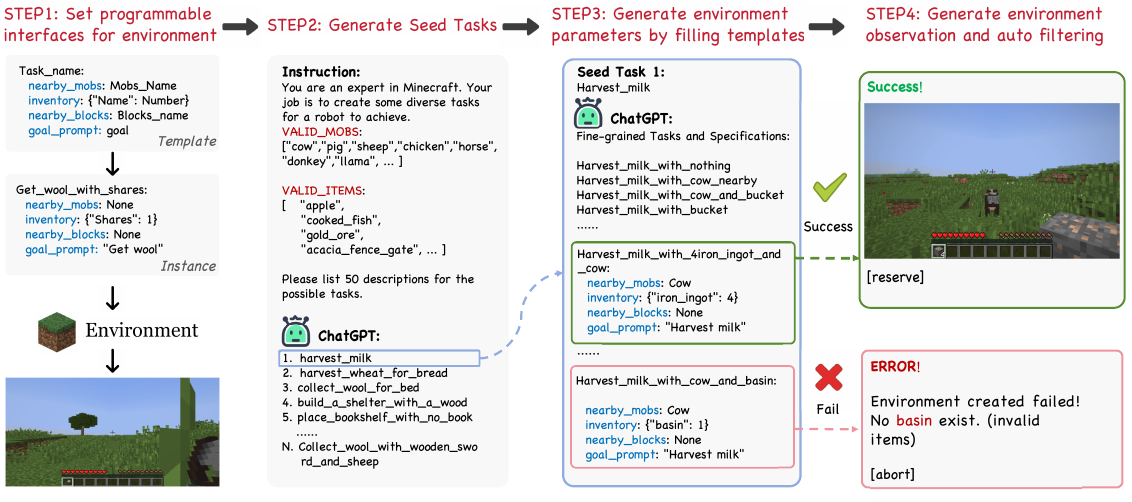}
    \caption{Pipeline of the Embodied Instruction Evolution method.
    }
    \label{fig:illustrative-example}
\end{figure*}

The annotation of PCA-Bench examples is a labor-intensive task. As illustrated in Figure~\ref{fig:illustrative-example}, we introduce Embodied Instruction Evolution (EIE), a method for automatically augmenting examples in the PCA-Bench format using Large Language Models, such as ChatGPT. This process involves four key steps:

\textbf{1) Setup of Programmable Interface:} Establish a programmable interface with a corresponding template, ensuring that observations in the embodied environment can be generated based on specific parameters.

\textbf{2) Generation of Seed Tasks:} Create initial seed tasks for each environment. These tasks are representative of the general challenges an agent might encounter. We provide ChatGPT with sample tasks and enable it to generate additional seed tasks.

\textbf{3) Task Specification and Template Filling:} For each seed task, we instruct ChatGPT to break down the task into multiple subtasks, following its event topology graph (as seen in Figure~\ref{fig:topology}). This approach mimics the multi-step decision-making process. After determining the subtask names, we use the LLM to populate the environment parameter templates created in Step 1 for each subtask.

\textbf{4) Observation Generation and Filtering:} Generate observations for the environment and implement an automatic process to filter out invalid instances. The filled templates may contain errors, such as incorrect creature names or impossible items, leading to errors during environment creation. When such errors occur, the affected templates are automatically filtered out. For domains without programmable environments (autonomous driving), step 1 and step 4 are not needed, we collect real traffic images and utilize GPT4-Vision to generate seed task based on the image content. 

EIE leverages the capabilities of Large Language Models to reduce manual labor and improve the diversity and scalability of PCA-Bench.

\section{Experiments}

\begin{table*}[t]
    \centering
    \resizebox{\textwidth}{!}{
    \begin{tabular}{l|c|ccc|ccc|ccc|ccc}
    \toprule
        \multirow{2}{*}{Model} & \multirow{2}{*}{Size}  & \multicolumn{3}{c|}{Traffic} & \multicolumn{3}{c|}{Domestic}  & \multicolumn{3}{c|}{Game} & \multicolumn{3}{c}{Average} \\ 
        ~ & ~  & P & C & A & P & C & A & P & C & A &P & C & A \\
        \midrule



        MiniGPT4~\citep{zhu2023minigpt}\color{red}{$^\dagger$}  & 7B & 0.45 & 0.37 & 0.48 & 0.81 & 0.38 & 0.38 & 0.38 & 0.14 & 0.27 & 0.55 & 0.30 & 0.38  \\

        LLaVA1.5~\citep{liu2023llava}\color{red}{$^\dagger$} & 7B  &0.44 & 0.44 & 0.53 & 0.92 & 0.48 & 0.44 & 0.8 & 0.35 & 0.39 & 0.72 & 0.42 & 0.45  \\

        Qwen-VL-Chat~\citep{Bai2023QwenVLAF}\color{red}{$^\dagger$}& 7B  &0.53 & 0.36 & 0.62 & 0.77 & 0.41 & 0.44 & 0.39 & 0.18 & 0.25 & 0.56 & 0.33 & 0.44  \\

        MiniGPT4~\citep{zhu2023minigpt}\color{red}{$^\dagger$}  & 13B & 0.41 & 0.37 & 0.5 & 0.85 & 0.35 & 0.33 & 0.41 & 0.22 & 0.33 & 0.56 & 0.31 & 0.39   \\

        InstructBLIP~\citep{Dai2023InstructBLIPTG}\color{red}{$^\dagger$}  & 13B & 0.36 & 0.41 & 0.42 & 0.90 & 0.44 & 0.39 & 0.33 & 0.25 & 0.24 & 0.53 & 0.37 &0.35  \\ 
        
    MMICL~\citep{zhao2023mmicl}\color{red}{$^\dagger$}  & 13B & 0.31 & 0.49 & 0.47 & 0.81 & 0.3 & 0.33 & 0.41 & 0.18 & 0.27 & 0.51 & 0.32 & 0.36  \\ 

        SPHINX-v1~\citep{lin2023sphinx}\color{red}{$^\dagger$} & 13B & 0.46 & 0.48 & 0.61 & \underline{0.95} & 0.55 & 0.31 & 0.71 & 0.35 & 0.43 & 0.71 & 0.46 & 0.45  \\

        LLaVA1.5~\citep{liu2023llava}\color{red}{$^\dagger$}& 13B  & 0.49 & \underline{0.56} & 0.61 & \underline{0.95} & \underline{0.62} & \underline{0.46} & \underline{0.74} & 0.45 & \underline{0.51} & \underline{0.73} & \underline{0.54} & \underline{0.53}  \\ 

        Qwen-VL-Chat-PLUS~\citep{Bai2023QwenVLAF}$^\ddagger$& UNK &  \underline{0.57} & \underline{0.56} & \underline{0.65} & 0.86 & 0.44 & 0.43 & 0.68 & \underline{0.47} & 0.49 & 0.70 & 0.49 & 0.52  \\

        GPT-4V~\citep{gpt4v}$^\ddagger$& UNK  & \textbf{0.73} & \textbf{0.72} & \textbf{0.74} & \textbf{0.96} & \textbf{0.66} & \textbf{0.62} & \textbf{0.88} & \textbf{0.72} & \textbf{0.69} & \textbf{0.86} & \textbf{0.7} & \textbf{0.68} \\ 

        \bottomrule
    \end{tabular}}
    \caption{Zero Shot results on the full test set of PCA-Bench. Highest scores in each line are \textbf{bold} while second highest scores are \underline{underlined}. Models with \color{red}{$\dagger$} \color{black} are fully open-source. Models with $\ddagger$ only provide API to access. P, C, and A represent Perception, Cognition, and Action Scores, respectively. }
    \label{tab:main}
\end{table*}

\begin{figure*}[t]
    \centering
    \includegraphics[width=\textwidth]{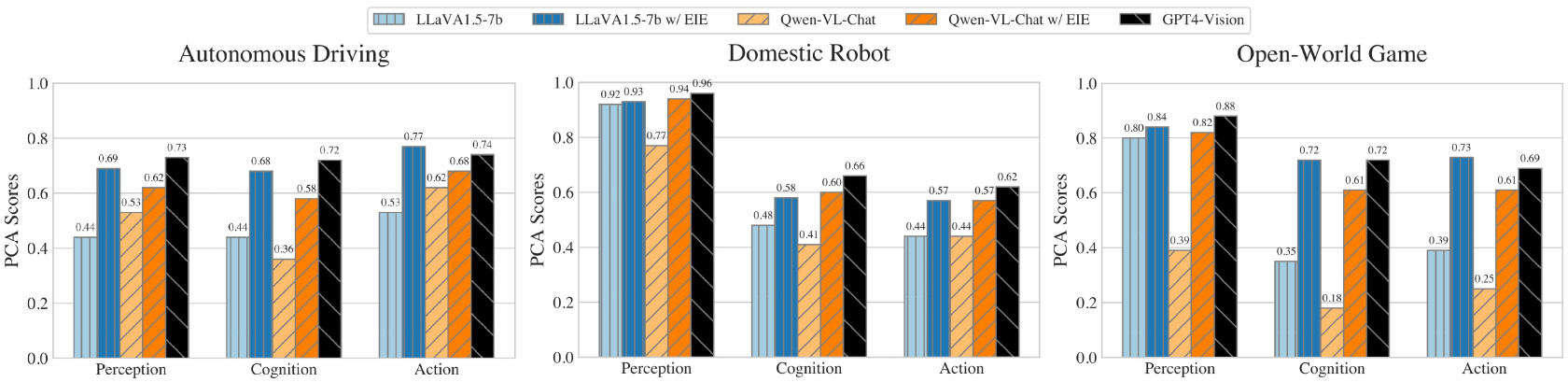}
    \caption{Performance comparsion between models' zero-shot results and models' finetuned results with the data generated by Embodied-Instruct-Evolution (EIE) method. EIE improves the performance on all domains for both LLaVA1.5-7b and Qwen-VL-Chat models. Results of LLavA1.5-13B and MMICL are in Figure~\ref{fig:sft-results-mmicl} from appendix.}
    \label{fig:sft-results}
\end{figure*}

\subsection{Tracks}
\paragraph{Zero Shot End-to-End.}

The test set of PCA-Bench serves as an effective tool for comparing the embodied decision-making and cross-modal reasoning capabilities of various Multimodal Language Learning Models (MLLMs). In this evaluation, the same images and prompts are provided to each model under test. Additionally, to address the challenge of perceiving certain non-visual information from images, details such as ``items in hand'' and ``items in inventory'', particularly relevant in domestic and gaming domains, are directly included in the question prompts. 

In our analysis, we benchmark the performance of the most recently open-sourced models, including LLaVA1.5 and Qwen-VL-Chat, as well as the API-only GPT4-V model. All models are evaluated using their default inference configurations to ensure a fair and standardized comparison. 

\paragraph{Finetuning with EIE.} In this track, we extend the capabilities of open-source MLLMs by fine-tuning them with the training set generated through our Embodied Instruction Evolution (EIE) method.  After the fine-tuning process, these trained models are subjected to the test set of PCA-Bench. We finetune the LLaVA-7b/13b, MMICL and Qwen-VL-Chat models on the training set for 5 epochs. The training details are in Appendix~\ref{app:training_details}.

\paragraph{Zero Shot Modality Conversion.}

In this track, we introduce and compare a new baseline, termed HOLMES, which utilizes LLM without multimodal perception capabilities. Instead, HOLMES relies on modality conversion APIs for embodied decision-making processes. Within the HOLMES framework, the LLM must continuously invoke various APIs, retrieving and processing return information about the environment. The HOLMES method is illustrated in Figure~\ref{fig:illustrative-example-holmes} from Appendix. 

We evaluate two LLMs in this track: ChatGPT-3.5-Turbo and GPT-4-0613, comparing their performances against the advanced GPT-4-Vision. Implementation details of the HOLMES framework and the APIs are provided in Appendix~\ref{app:api_des}.

\subsection{Evaluation and Metrics}

We use our PCA-Eval evaluation tool proposed in Section~\ref{sec:pca-tool} to automatically assess the output of different models through three lenses: perception (P-Score), cognition (C-Score), and action (A-Score).

\subsection{Main Results}

\begin{table*}[t]
    \centering
    \resizebox{\textwidth}{!}{
    \begin{tabular}{c|c|ccc|ccc|ccc|ccc}
    \toprule
        \multirow{2}{*}{Method} & \multirow{2}{*}{Model}  & \multicolumn{3}{c|}{Traffic} & \multicolumn{3}{c|}{Domestic}  & \multicolumn{3}{c|}{Game} & \multicolumn{3}{c}{Average} \\ 
        ~ & ~  & P & C & A & P & C & A & P & C & A &P & C & A \\
        \midrule
        \multirow{1}{*}{End-to-End}



        ~ & GPT-4V  & 0.75 & 0.73 & 0.78 & 0.81 & \textbf{0.69} & \textbf{0.67} & \textbf{0.95} & \textbf{0.79} & \textbf{0.77} &0.84 & \textbf{0.74} & \textbf{0.74} \\ 
        \midrule
        
        \multirow{2}{*}{HOLMES} & ChatGPT  & 0.75 & 0.68 & 0.66 & \textbf{0.88} & 0.52 & 0.50 & 0.78 & 0.40 & 0.36 & 0.80 & 0.53 & 0.51 \\ 
        ~ & GPT4  & \textbf{0.87} & \textbf{0.82} & \textbf{0.82} & 0.85 & 0.61 & 0.56 & 0.91 & 0.77 & 0.74 & \textbf{0.88} & 0.73 & 0.71 \\ 
        \bottomrule
    \end{tabular}}
    \caption{Comparison between End-to-End (MLLM) and HOLMES (LLM+API) methods on a subset of PCA-Bench with API annotation.}
    \label{tab:holmes}
\end{table*}



\paragraph{Zero Shot Results.}
The results of the zero-shot end-to-end track are shown in Table~\ref{tab:main}. Among all MLLMs, GPT4-V, outperforms existing open-source models by achieving the highest scores of 0.86, 0.7, and 0.68 in the perception, cognition, and action dimensions respectively. 
This performance represents a 15\% action score improvement over its strongest open-source counterpart, LLaVA1.5-13B. 
The impressive performance of GPT4-V is primarily attributed to its exceptional ability to perceive visual information across different domains and the world knowledge in the language model, particularly in the challenging game domain.

\paragraph{Impact of Finetuning with EIE.}
The results of the fine-tuning track are illustrated in Figure~\ref{fig:sft-results}. Our EIE method has been found to significantly enhance the general decision-making abilities of various models, encompassing perception, cognition, and action. Notably, it has led to an average increase of 0.24 and 0.19 in action scores for the LLaVA1.5-7b and Qwen-VL-Chat models, respectively. Results for LLaVA1.5-13b and MMICL are illustrated in Figure~\ref{fig:sft-results-mmicl}, also showing improved performance when trained with EIE. We note that there exist reasoning or perception errors in some of the generated sample due to the hallucination problem of LLM generated content, however they do not influence the overall performance. In some cases, these sub-scores have matched or even surpassed those of the GPT4-V model, demonstrating the potential of the EIE to scale up and apply to different environments.

\paragraph{Comparison Between End-to-End and Modality Conversion Method}

In the zero-shot modality conversion track, we conduct an analysis and comparison of the outputs generated by the End2End method with GPT4-V, as well as the HOLMES method with GPT4 and ChatGPT-3.5 in Table~\ref{tab:holmes}.

The results show that the HOLMES system based on GPT4 achieves 0.71 Action Score, which is on par with GPT4-V's performance (0.74). This indicates that, overall, the HOLMES system is able to accurately understand the task goal, split the larger goal into multiple smaller steps, and correctly invoke the relevant APIs to accomplish each step. 
Specifically, the HOLMES system based on GPT4 can recognize the key concepts in a task, and perceive the state and environment of these concepts through the results returned by APIs. Consequently, the system achieves an average Perception Score of 0.88, which even outperforms GPT4-V's 0.84. However, compared to End2End methods, HOLMES relies on multi-step reasoning for the final decision, in which reasoning errors tend to accumulate, and thus achieves a lower Cognition Score in both Domestic and Game domains. 

On the other hand, we also find that the End2End method effectively mitigates information loss during the modality conversion process. As illustrated in Figure~\ref{fig:compare} from Appendix, an image depicts a road with several nearby cars. GPT4-V is capable of discerning that the street is not crowded, thereby suggesting that the driver can continue driving. 

Conversely, GPT4-HOLMES, while being aware of the number of cars, lacks information about their spatial relation, leading it to recommend slowing down because of the existence of 14 cars. This suggests that the End2End method is superior in perceiving certain visual features that are not captured by the APIs. Conversely, some specialized APIs, such as traffic sign detection, outperform GPT4-V in tasks like traffic sign detection, as they are specifically trained for this task. This could enable the HOLMES method to gather more accurate information than the End2End model.
\section{Discussion}

\subsection{Strong LLMs are Good Error Locators.}
\label{llm-eval}

As shown in Table~\ref{tab:llm-evals}, we compare the scoring kappa coefficients with human assessments for different LLMs. We randomly select 300 model outputs equally from different domains and ask 3 human experts to give perception, cognition, and action scores. The final result is based on the majority of three annotators. The result underscores a significant agreement between GPT-4 scoring and human evaluation results. This is substantiated by Cohen-Kappa coefficients of 0.71, 0.82, and 0.94 for perception, cognition, and action evaluations.

\begin{table}[t]
\centering
\resizebox{0.4\textwidth}{!}{
\begin{tabular}{c|ccc}
\toprule
\multirow{2}{*}{Evaluator Model} & \multicolumn{3}{c}{Kappa Coefficients}\\
~ & P & C & A \\
\midrule
GPT4$^\dagger$ &\textbf{0.71}&\textbf{0.82}&\textbf{0.94}\\
Qwen1.5-72B-Chat\color{red}{$^\dagger$}&0.30&0.49&0.60\\
Qwen1.5-14B-Chat\color{red}{$^\dagger$} &0.16&0.24&0.16\\
Qwen1.5-7B-Chat\color{red}{$^\dagger$} &0.20&0.11&0.06\\
\bottomrule
\end{tabular}}

\caption{Comparison of Open\color{red}{$^\dagger$} \color{black} and Close$^\dagger$ LLMs as Evaluators. Kappa coefficients of Qwens increase when the model scales up.}
\label{tab:llm-evals}
\end{table}

We also compare open models as evaluators. We choose one of the best open LLMs,  Qwen1.5\footnote{\url{https://huggingface.co/collections/Qwen}} series from 7B, 14B to 72B version. \textit{Currently open LLMs tend to give wrongly high judgments in all sub-scores.} Although currently trailing behind GPT-4 in performance, we anticipate that with targeted training focused on error identification and enhancements in the overall capabilities of open LLMs, these models will become more effective evaluation tools compared to closed models. This is primarily due to the reproducible and transparent nature of open models, which offer significant advantages in the development of evaluation tools.

\subsection{Genuine PCA Score}
\label{genuine_pca}
PCA-Eval could pinpoint cases where the MLLM gets the correct answer by a fluke where perception or cognition score is 0 but the action score is 1. It explains why for some models, the action score is higher than perception and cognition scores. For instance, a model might opt for a conservative action, such as slowing down, even without accurately recognizing snowy weather in the image, resulting in a fluky correct action. In another scenario, if the model exhibits a preference for a specific choice index, it will attain a high action score provided that the evaluation dataset contains a substantial number of correct choices matching the preferred index, a phenomenon attributable to the positional biases inherent in both the model and the dataset. To overcome the mentioned bias when evaluating the genuine ability of MLLM, we propose a new metric \textbf{Genuine PCA Score}. It is equal to one if the perception, cogntion and action scores are all 1 for one model's response to a question. We find that for all models, there exists significant gap (>10\%)  between the action score and genuine PCA score in average, revealing that relying on single metric such as choice accuracy is very problematic when conducting model evaluation. In our \href{https://docs.qq.com/sheet/DVUd4WUpGRHRqUnNV?tab=mqeh4c}{online leaderboard}, both average action score and average genuine PCA score are considered when ranking the candidate models.

\subsection{Alignment between Agent Decisions and Human Values}






We have observed instances where the decisions made by the agent contradict human values. Consider the scenario depicted in Figure~\ref{fig:alignment} from Appendix. The image illustrates a crosswalk without pedestrians. The appropriate response would be slowing down, as caution is paramount when approaching a crosswalk, regardless of the presence or absence of pedestrians. However, upon processing the information that the crosswalk is empty, ChatGPT suggests that maintaining the current speed is the optimal action, arguing that the absence of pedestrians eliminates the need to slow down. The rationale provided by ChatGPT is logical, yet it does not align with human values.

\section{Related Work}

\paragraph{MLLM Benchmark.}
In recent times, there have been several benchmarks built for evaluating MLLMs, such as MMBench, MME, Seed-Bench, POPE~\citep{liu2023mmbench,fu2023mme, Li2023SEEDBenchBM, POPE} that assess MLLMs performance from multiple fine-grained dimensions. Visit-Bench, LVLM-eHub, M3IT~\citep{bitton2023visitbench,xu2023lvlm,li2023m3it} focus on the general instruction following ability. General VQA tasks like OKVQA, VQAv2, Vizwiz, ScienceQA, VSR and IconQA~\citep{marino2019okvqa, Agrawal2015VQAVQ, Gurari2018VizWizGC, scienceqa, Liu2022VisualSR, lu2021iconqa} focus on visual understanding. MMMU, MathVista, LLaVA-benchmark and MM-Vet~\citep{yue2023mmmu, Lu2023MathVistaEM, llava, Yu2023MMVetEL} require abilities from the vision part and specific knowledge in the language part.
A lack of error localization techniques beyond accuracy assessments is among current benchmarks. This complicates identifying which part of the MLLM malfunctioned when making mistakes. Unlike prior work, PCA-Bench is more relevant to evaluate MLLMs' ability to utilize integrated abilities to solve one task and make explainable decisions via error localization.

\paragraph{LLM Agent and Embodied Decision Making.}

Using LLMs to empower the AI agents~\citep{xi2023rise,liu2023training,park2023generative,Wang2023DescribeEP} becomes more and more promising. 
Specifically, we can employ LLMs to enhance the decision making ability of the agents~\citep{nakano2022webgpt,yao2022react,li2023apibank,song2023restgpt,li2023camel}, expanding their perception and action space through strategies like tool utilization~\citep{schick2023toolformer,qin2023tool,lu2023chameleon}. This line of research divides the entire decision-making process into two phases: (1) information seeking, usually involving MLLMs to verbalize the current status of AI agents in the vision-based environment with natural language; (2) reasoning and planning with text-based LLMs to decide what the AI agent should do in the next step with textual clues. Although LLM-based agents demonstrate reasoning and planning abilities through techniques like Chain of Thought or problem decomposition~\citep{wei2023chainofthought,yao2023retroformer,kojima2022large}, they inherently lack visual perception, and are limited to the discrete textual content. Therefore, integrating multimodal information can offer agents a broader context and a more precise understanding, such as PaLM-E~\citep{driess2023palme}, enhancing their environmental perception. However, there is still large gap deploying MLLM in various embodied environments due to the lack of appropriate benchmark and interface linking those two domains while PCA-Bench is an attempt towards that goal.

\section{Conclusion}

 

In this paper, we introduce PCA-Bench, a multimodal benchmark designed to assess the integrated decision-making capabilities of MLLMs. This benchmark features PCA-EVAL, a novel fine-grained automatic evaluation tool that diagnoses decision making processes from three critical perspectives: perception, cognition, and action. To enhance the decision making ability from data perspective, we propose the Embodied Instruction Evolution method to automatically synthesize instruction examples from different environments, which has been proven effective in our main experiments. We believe that powerful MLLMs pave a new and promising way toward decision making in embodied environments and we hope PCA-Bench could serve as a good benchmark in evaluation and error localization for MLLMs' development.

\section{Limitations}


The current scope of PCA-Bench is confined to merely three domains in static environments. One of our future works aims to broaden this scope to encompass more domains and dynamic embodied environments where MLLMs could keep getting feedback, which is closer to real embodied AI scenarios. We do not apply different inference enhancement methods like In-Context Learning and Reflection in the decision making process of MLLMs. We just use the simplest prompting method and leave the exploration of a better cross-modal Chain-of-Thought method for future studies. 
Currently, PCA-Eval shows the best consistency with human evaluators when using powerful close LLM GPT4, which would bring additional cost to the user of PCA-Eval. We plan to develop and release an open error locator for error localization in the benchmark in the future.

\bibliography{custom}

\appendix

\clearpage

\section{Examples of PCA-Bench}
\label{app:example_pca_eval}
\subsection{Data Distribution}

\begin{figure}[h]
    \centering
    \includegraphics[width=0.35\textwidth]{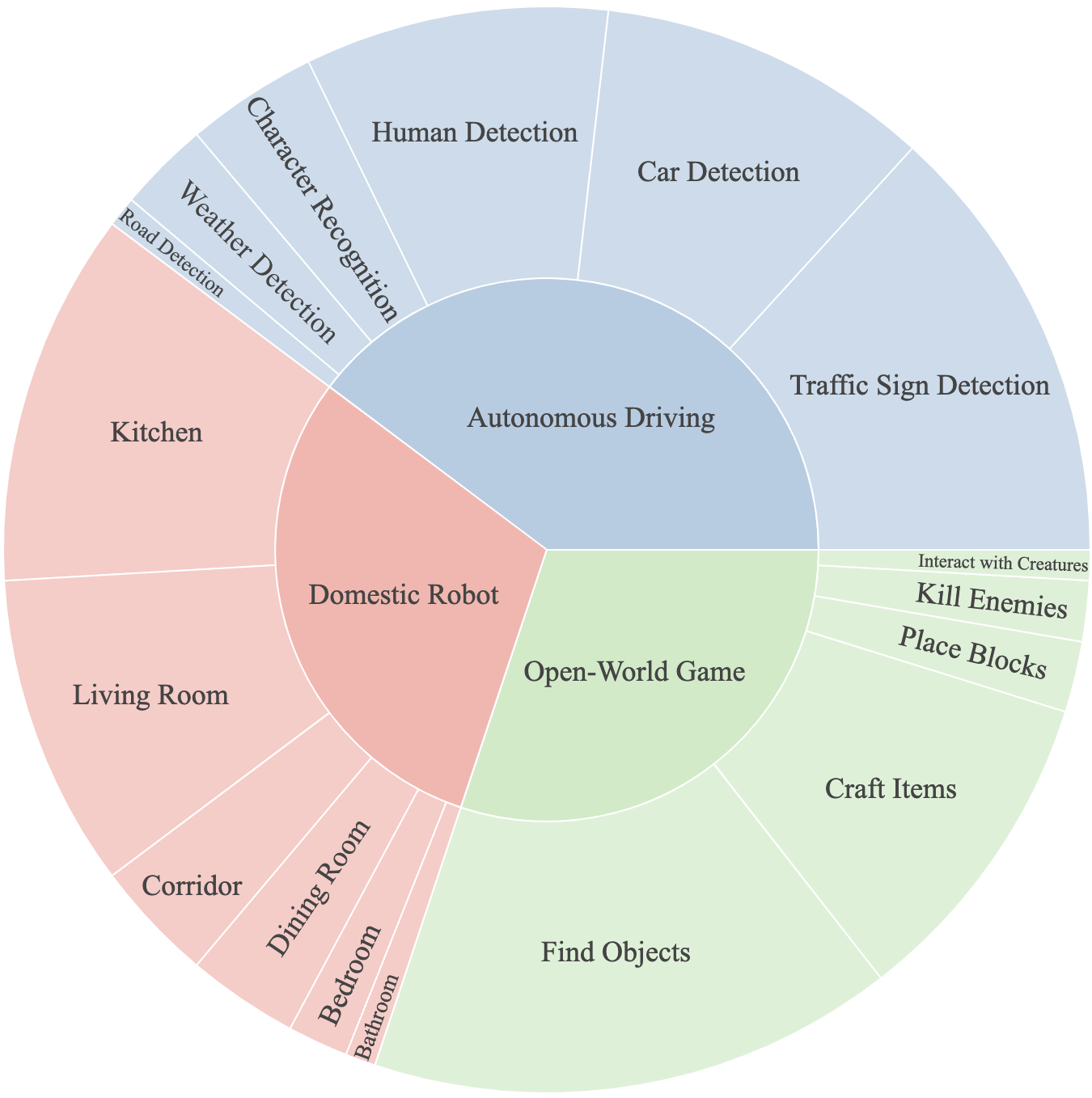}
    \caption{Domain and required ability distribution of PCA-Bench.
    }
    \label{fig:sun-figure2}
\end{figure}

\vspace{-8pt}

The PCA-Bench's data distribution across various domains is outlined in Figure~\ref{fig:sun-figure2}. 
For the Autonomous Driving domain, instances are grouped by their respective task types. In the Domestic Robot domain, instances are grouped by their locations. In the Open-World Game domain, instances are grouped by the tasks they aim to accomplish.






\section{Human Annotation Pipelines}
\label{app:human_anno_pipelines}

The annotation process consists of two stages: (1) Dataset Annotation, and (2) Dataset Refinement. During the initial stage, three annotators are assigned to each domain, adhering strictly to the respective annotation guidelines. They first pinpoint the source images from each domain that are informative and meaningful so that they can write questions for each image. The annotators have the responsibility to ensure every question has only one correct answer and accurate rationales. In the subsequent stage, annotators are instructed to scrutinize the output actions and rationales presented by ChatGPT and check the annotations. This process aims to address the challenge of multiple correct answers, as ChatGPT can furnish comprehensive explanations for its actions. These explanations assist annotators in assessing the acceptability of ChatGPT's response, particularly when it deviates from the established ground truth answer. This enables annotators to refine annotations to ensure the presence of a single correct answer. 


\subsection{PCA-Bench Examples}
We list three examples of each domain from PCA-Bench, as shown in Figure~\ref{fig:pca-examples-traffic},~\ref{fig:pca-examples-domestic}, and~\ref{fig:pca-examples-game}.

\section{Zero Shot Modality Conversion: HOLMES}

To optimize the evaluation process of HOLMES\footnote{Originally proposed in an early version of this paper~\citep{chen2023endtoend}} method, we pre-execute all relevant APIs for each instance within a selected subset of 300 instances from the PCA-Bench test set, recording the results for individual instances. This method enables immediate access to specific API results, eliminating the need to rerun the model for each evaluation instance.

\begin{figure*}[t]
    \centering
    \includegraphics[width=\textwidth]{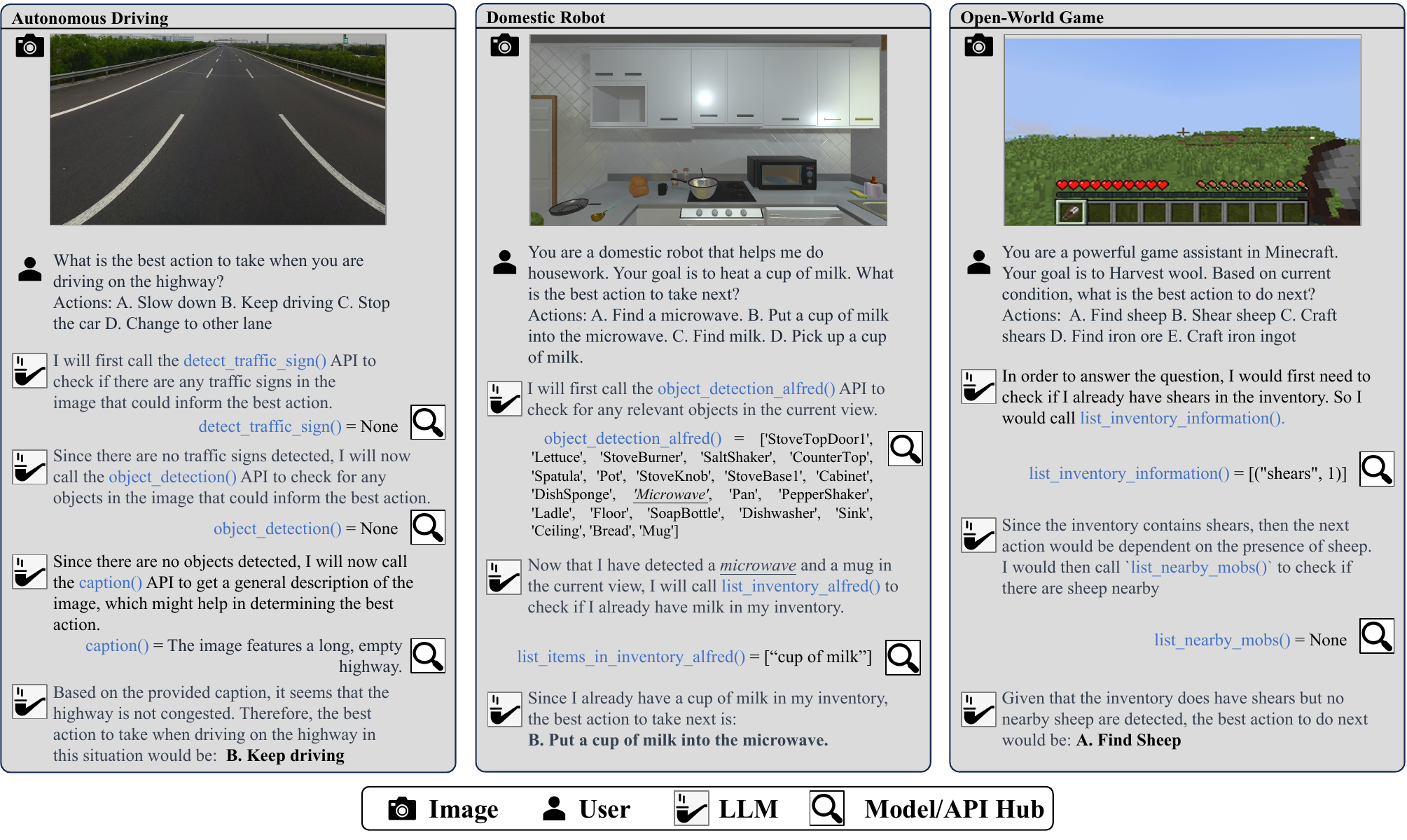}
    \caption{Three examples of HOLMES solving questions from different domains of PCA-Bench.
    }
    \label{fig:illustrative-example-holmes}
\end{figure*}

\label{app:api_des}
\paragraph{Traffic Domain.} Below is the API description for the traffic domain.
\begin{lstlisting}[language=Python]
# API Description for Traffic Domain:
def detect_traffic_sign():
    """
    Detects traffic signs in the image.
    :return: list of detected traffic signs and coordinates, e.g. ['stop','max speed limit']
    """
    pass

def object_detection():
    """
    Detects objects in the image.
    :return: dict of detected objects and number of the objects, e.g. {'car':10, 'person':1}
    """
    pass

def ocr():
    """
    Performs OCR on the image.
    :return: list of detected text, e.g. ['Changjiang road', 'Right lane closure']
    """
    pass

def image_caption():
    """
    Generates a caption for the image.
    :return: caption, e.g. 'A red car driving down the street'
    """
    pass

def weather_detection():
    """
    Detect current weather.
    :return: weather, e.g. 'rainy' or 'clear'
    """
    pass
\end{lstlisting}

\textbullet\ \textit{detect\_traffic\_sign()}: The detection of road traffic signs model utilize YOLO~\citep{YOLO} which trained on the Tsinghua-Tencent 100K dataset~\citep{Zhe_2016_CVPR_traffic_sign}. TT100K comprises 100,000 images encompassing 30,000 instances of traffic signs. The end-to-end YOLO enables simultaneous detection and classification of traffic signs.

\textbullet\ \textit{object\_detection()}: Objects demanding attention during vehicle operation primarily encompass cars, pedestrians, and bicycles. A surfeit of vehicles can lead to traffic congestion, while the presence of pedestrians or bicycles ahead necessitates cars to decelerate and proceed cautiously. Hence, the \textit{object\_detection()} API predominantly identifies three key object categories: cars, pedestrians, and bicycles. We utilize PMOP~\citep{ren2023pomp}, a model trained on vision-language models through the prompt pre-training method, which enables the detection and counting of the three mentioned objectives by modifying specific class names.

\textbullet\ \textit{ocr()}: We employ PaddleOCR\footnote{\url{https://github.com/PaddlePaddle/PaddleOCR/tree/release/2.7}} to extract textual information from images, providing crucial road data for real-time navigation.

\textbullet\ \textit{image\_caption()}: To initially streamline the road information within the image, we employ the BLIP2-flan-t5-xl to generate an initial caption for the picture. This caption, derived from basic image data, is then utilized as input for the model to facilitate decision-making.

\textbullet\ \textit{weather\_detection()}: Weather detection leverages a pre-trained ResNet50 model\footnote{\url{https://github.com/mengxianglong123/weather-recognition}}, derived from a dataset of more than 70,000 weather records. This model extracts weather information from provided images to inform decision-making.

\paragraph{Domestic Robot Domain.} Below is the API description for the Domestic Robot domain.
\begin{lstlisting}[language=Python]
#API Description for Domestic Robot Domain
def object\_detection():
    """
    Detects objects in current view, which you don't need do find.
    :return: list of detected objects, e.g. ['chair','table']
    """
    pass

def list_items_in_hands():
    """
    Lists items in your hand, which you don't need to pick up
    :return: list of items in hand, e.g. ['coffee cup','milk']
    """
    pass
\end{lstlisting}

\paragraph{Game Domain.} Below is the API description for the Game domain (Minedojo).
\begin{lstlisting}[language=Python]
#API Description for Game Domain
def list_nearby_mobs_in_minecraft():
    """
    Lists nearby mobs in Minecraft.
    :return: list of nearby mobs, e.g. ['creeper', 'pig']
    """
    pass

def list_inventory_information():
    """
    Lists inventory information of the player in Minecraft.
    :return: list of inventory information with number, e.g. [('diamond', 64), ('iron', 32)]
    """
    pass
\end{lstlisting}

Note that within the Domestic Robot Domain and Game Domain, APIs can be directly accessed within the virtual environment, allowing for the perception of the surrounding objects and the current picture context.

\begin{figure}[htbp]

  \begin{center}
    \includegraphics[width=0.47\textwidth]{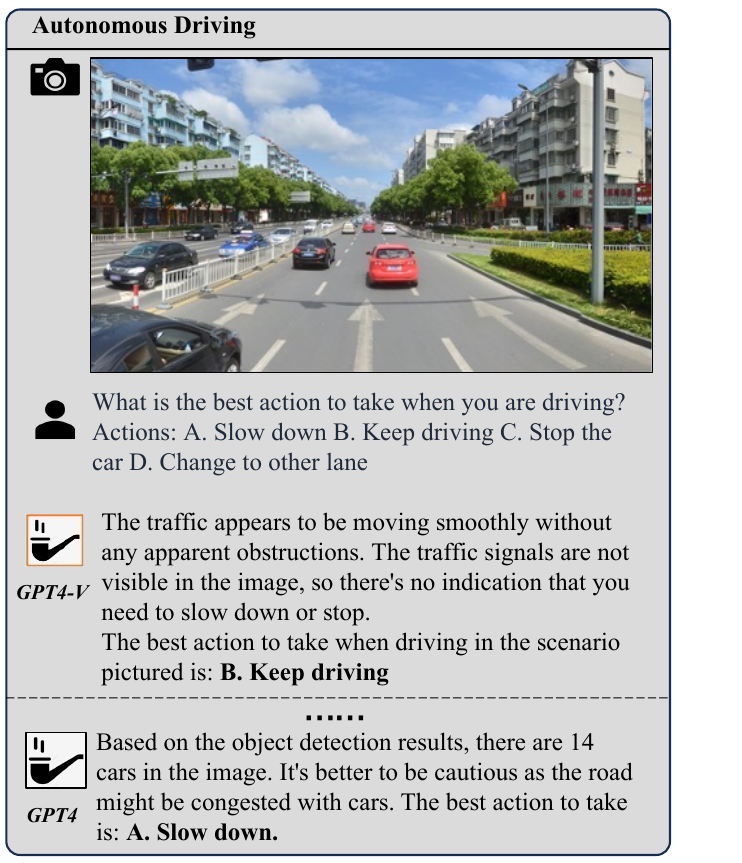}
  \end{center}

  \caption{A Comparison between GPT4-V and GPT4-HOLMES}
  \label{fig:compare}

\end{figure}

\begin{figure}[htbp]

  \begin{center}
    \includegraphics[width=0.47\textwidth]{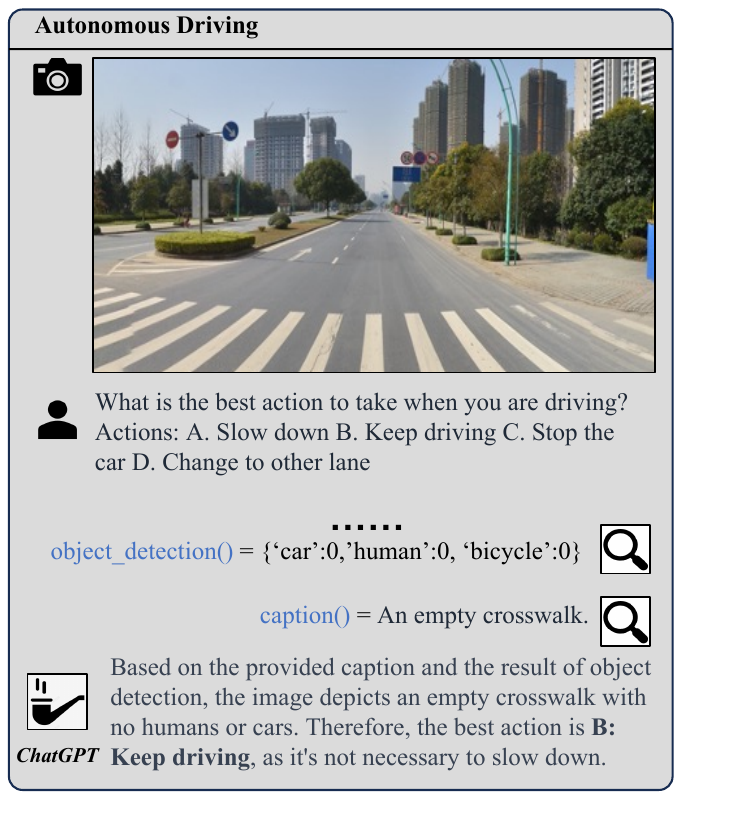}
  \end{center}

  \caption{An case showing the value mis-alignment between of agent and human's decision. }
  \label{fig:alignment}

\end{figure}

\begin{figure*}[t!]
    \centering
    \includegraphics[width=0.93\textwidth]{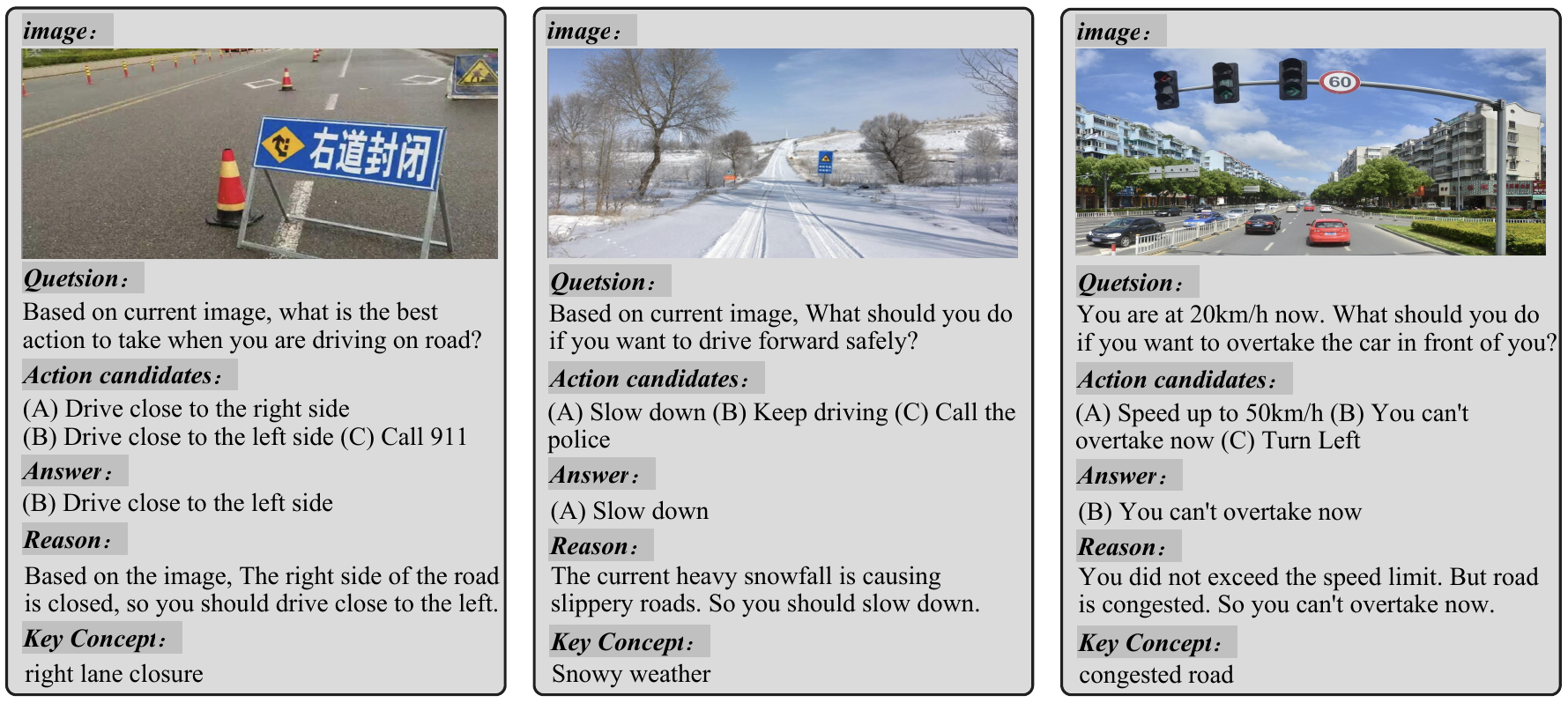}
    \caption{Three examples of PCA-Bench in the autonomous driving domain.
    }
    \label{fig:pca-examples-traffic}
\end{figure*}
\vspace{-8pt}
\begin{figure*}[t!]
    \centering
    \includegraphics[width=0.93\textwidth]{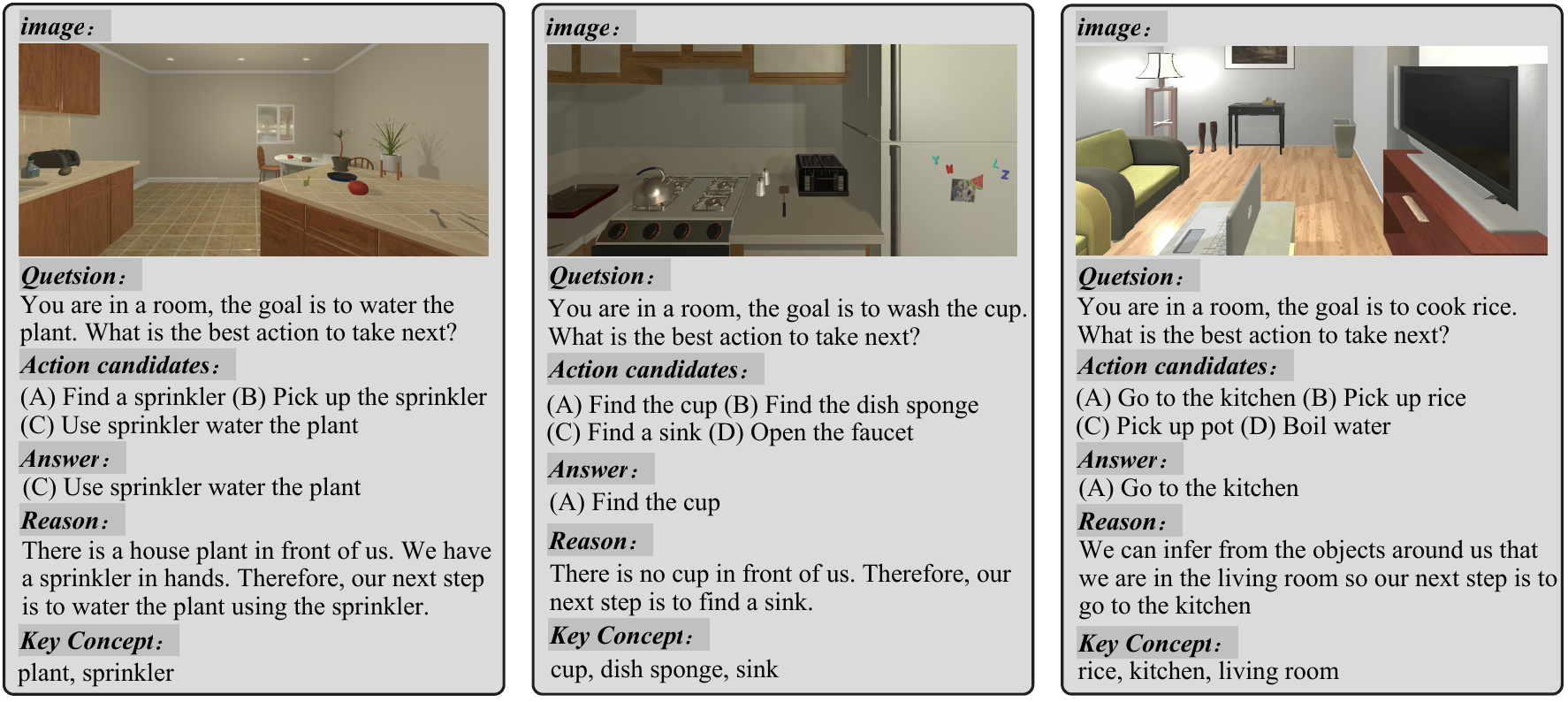}
    \caption{Three examples of PCA-Bench in the domestic robot domain.
    }
    \label{fig:pca-examples-domestic}
\end{figure*}
\vspace{-8pt}
\begin{figure*}[t!]
    \centering
    \includegraphics[width=0.93\textwidth]{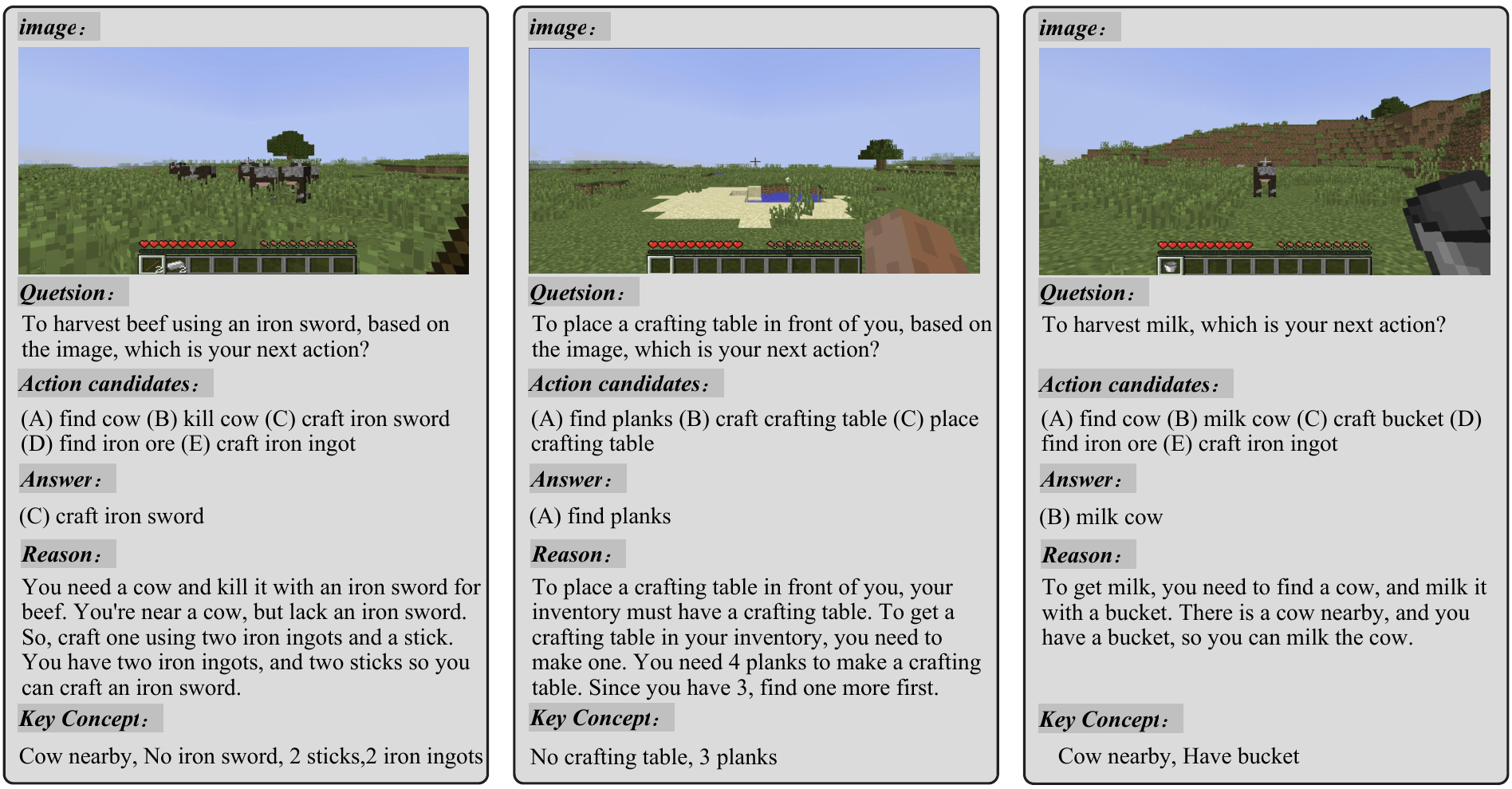}
    \caption{Three examples of PCA-Bench in the open-world game domain.
    }
    \label{fig:pca-examples-game}
\end{figure*}

\newpage
\newpage

\section{Automatic Evaluation}
\label{app:ae}

We utilize the template as shown in Table~\ref{tab:gpt4-eval} to query GPT-4, aiming to evaluate its responses and assign scores for perception,
cognition, and action. By feeding both the agent’s output and the ground truth answer to GPT-4, based on this
template, we can then extract the three distinct scores from the conclusion of GPT-4’s response.

\begin{table*}[ht!]

\begin{tcolorbox}

[Question]: \{question\}

[Action Choices]: \{actions\}

[Agent Answer]: \{model\_output\}

[Correct Action]: \{true\_action\}

[Key Concepts]: \{key\_concept\}

[Reference Reasoning Process]: \{reason\}

[System]

We would like you to access the agent's performance in the multimodal reasoning task about {domain}.
In this task, the agent is given an image, a [Question], and several candidate [Action Choices], and is asked to give an [Agent Answer] for the [Question].
The [Agent Answer] encapsulates the agent's perception of the image's [Key Concepts], the agent's cognition reasoning process and the final selected action.

We request you to give three types of scores for the agent's [Agent Answer] in comparison to the given [Key Concepts], [Reference Reasoning Process] and [Correct Action]:

1. action score: If the selected action in the [Agent Answer] matches that of the [Correct Action], the action score is 1; otherwise, it is 0.

2. perception score: This score evaluates the model's capability to perceive and interpret observations. It is contingent on whether the [Agent Answer] includes any of the [Key Concepts] of the instance. If it accurately describes any one of the [Key Concepts], the score is 1; otherwise, it is 0.

3. cognition score: This score gauges the model's ability to reason, comprehend, and make informed decisions based on perceived input data and world knowledge. If the reasoning process in the [Agent Answer] aligns with the [Reference Reasoning Process], the score is 1; otherwise, it is 0.

Please note that there are only scores of 0 and 1.

You should carefully compare the [Agent Answer] with the [Correct Action], [Key Concepts] and [Reference Reasoning Process] to give your assessment.

You need first to give your assessment evidence and then the scores. 

Your output MUST contain 6 lines with the following format:

action assessment evidence: (assessment evidence here)

action score: (score here)

perception assessment evidence: (assessment evidence here)

perception score: (score here)

cognition assessment evidence: (assessment evidence here)

cognition score: (score here)

\end{tcolorbox}
\caption{The template of querying GPT-4.}
\label{tab:gpt4-eval}
\end{table*}

\begin{figure*}[h]
    \centering
    \includegraphics[width=\textwidth]{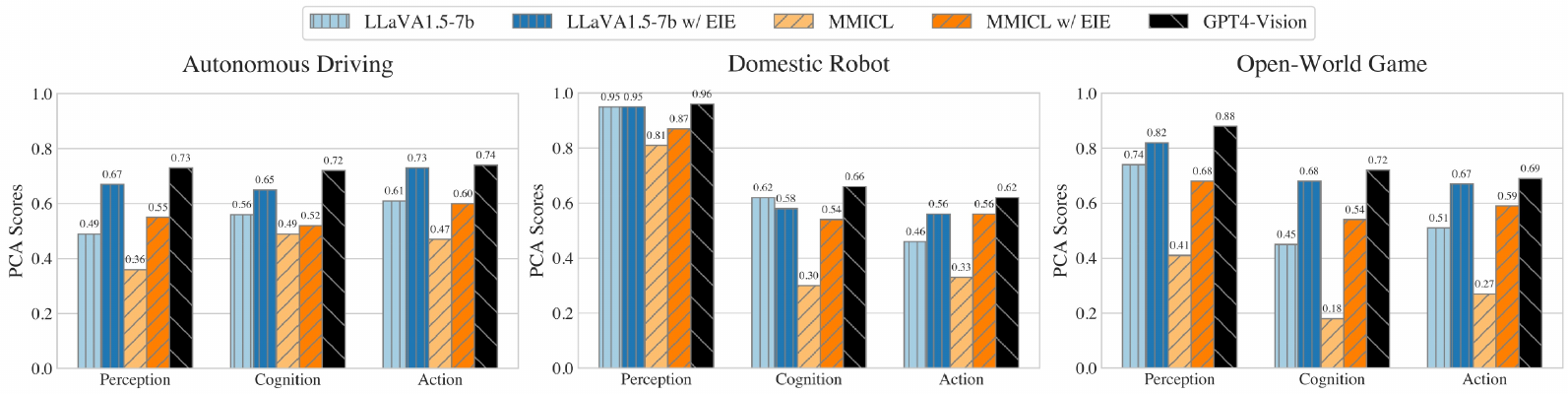}
    \caption{Performance comparsion between models' zero-shot results and models' finetuned results with the data generated by Embodied-Instruct-Evolution (EIE) method. EIE improves the performance on all domains for both LLaVA1.5-13b and MMICL models.}
    \label{fig:sft-results-mmicl}
\end{figure*}

\newpage

\section{Training Details}
\label{app:training_details}
Table \ref{tab:training-details} shows the specific parameters used for fine-tuning in different models. The PCA results on the three domains of PCA bench before and after fine-tuning different models are shown in Figure \ref{fig:sft-results-mmicl}.

\begin{table}[!h]
\centering
\resizebox{1\linewidth}{!}{
\begin{tabular}{lccc}
\toprule
\multicolumn{1}{c}{\textbf{Model}} & \multicolumn{1}{c}{\textbf{Parameter}} & \multicolumn{1}{c}{\textbf{Value}} &  \\
\midrule
& Learning Rate & 2e-4  \\
& Use Lora Finetuning? & Yes  \\
& Lora Rank & 8 \\
& Lora Alpha & 32 \\
Qwen-VL-Chat/LLaVA1.5-7/13b & Global Batchsize & 20 \\
& Weight Decay & 0 \\
& Train Epochs & 5 \\
& Lr Scheduler Type & Cosine \\
& Warmup Ratio & 0.03 \\
\midrule

& Learning Rate & 5e-4  \\
& Use Lora Finetuning? & No  \\
& Global Batchsize & 20 \\
MMICL & Weight Decay & 5e-4 \\
& Train Epochs & 5 \\
& Lr Scheduler Type & Linear \\
& Warmup Ratio & 0.2 \\

\bottomrule
\end{tabular}
}
\caption{Training details for different models with EIE.}
\label{tab:training-details}
\end{table}

\newpage
\clearpage

\section{Does Chain-of-Thought Finetuning Improve Cross-modal Reasoning?}

Unlike vanilla finetuning, which solely focuses on delivering direct answers, Chain-of-Thought Finetuning necessitates the model to first articulate its reasoning before presenting the answer. This approach has been demonstrated to be a highly effective instruction tuning paradigm for LLMs~\citep{chung2022scaling,kim2023cot}. We have incorporated this methodology in our previous finetuning experiments.

\begin{figure}[!t]
  \begin{center}
    \includegraphics[width=0.5\textwidth]{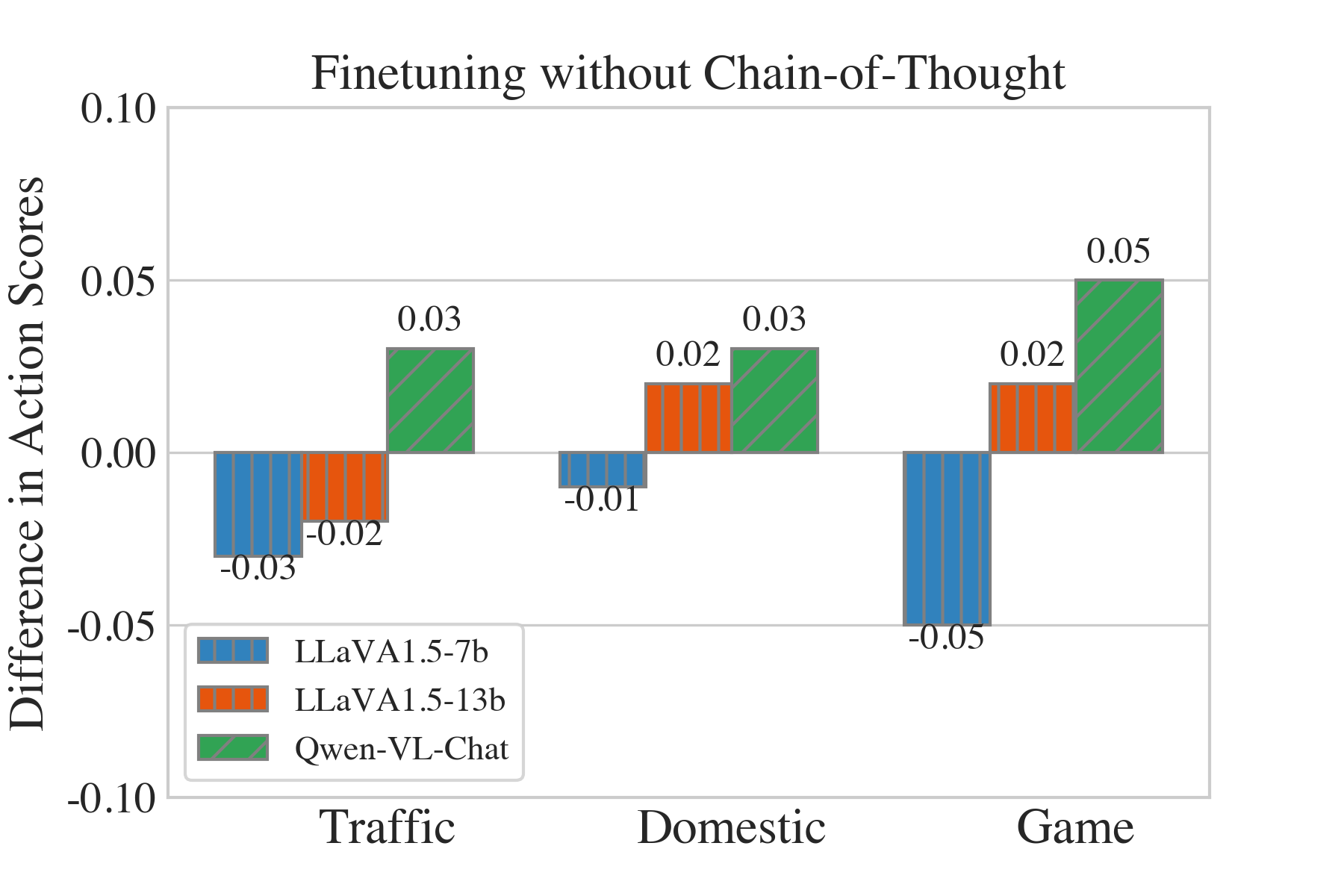}
  \end{center}
  \caption{Action scores changes when training without reasoning process for different models. The benefit of CoT finetuning is not consistent among models. Blue means difference of action score for LLaVA1.5-7b, Orange means difference of action score for LLaVA1.5-13b and green means difference of action score for Qwen-VL-Chat.}
  \label{fig:mmcot}
\end{figure}

To further evaluate its impact, we conducted an ablation study where the reasoning process was omitted from the target output during the training of MLLMs. We then assessed the variations in action scores on the test set. As depicted in Figure~\ref{fig:mmcot}, to our surprise, the figures suggest that Chain-of-Thought finetuning exerts a relatively minor influence when compared to conventional label finetuning. We have noticed that similar phenomena has been identified by~\citet{zhang2023multicot} that standard CoT finetuning does not work for MLLMs in their explorations.

We think there are three potential explanations: 1) Task Variation: Contrary to mathematics datasets like GSM8K, the current task doesn't require multi-step complex reasoning to arrive at the final answer and the automatic generated CoTs have noise. 2) Modality Discrepancy: The CoT capability, inherent in LLMs, is only moderately adjusted for visual input for current open-source MLLMs. This adaptation process could potentially impair the reasoning ability. 3) Short Cut in Pretraining: We think a deeper reason might lie in the short-cut during pretraining period of current open-source MLLMs, which are pretrained on simple image caption task in a large scale. Those captions are usually short and lose a lot of information about the original image. What's more important is that the reasoning ability of LLM is not utilized during the pretraining stage, which might hurt the reasoning ability of LLM during the SFT period. We defer to future research how to effectively harness the CoT capabilities of LLMs to enhance embodied decision-making processes.

\end{document}